\documentclass[conference,letterpaper]{IEEEtran}
\usepackage{fancyhdr}
\usepackage{graphicx}
\usepackage{amsmath}
\usepackage{amssymb}
\usepackage{algorithmic}
\usepackage{algorithm}
\usepackage{subfigure}
\usepackage{fancyhdr}

\newcommand{\bfy}{{\textbf{y}}}

\begin{document}

\title{Supervised cross-modal factor analysis for multiple modal data classification}

\author{
\IEEEauthorblockN{Jingbin Wang}
\IEEEauthorblockA{National Time Service Center\\
Chinese Academy of Sciences,
Xi' an 710600 , China\\
Graduate University of
Chinese Academy of Sciences\\
Beijing 100039, China\\
jingbinwang1@outlook.com}
\and
\IEEEauthorblockN{Yihua Zhou}
\IEEEauthorblockA{Department of Mechanical
Engineering and Mechanics\\
Lehigh University\\
Bethlehem, PA 18015, USA}
\and
\IEEEauthorblockN{Kanghong Duan}
\IEEEauthorblockA{
North China Sea Marine Technical Support\\
Center, State Oceanic Administration\\
Qingdao 266033, China}
\and
\IEEEauthorblockN{Jim Jing-Yan Wang}
\IEEEauthorblockA{
Computer, Electrical and Mathematical\\
Sciences and Engineering Division\\
King Abdullah University\\
of Science and Technology\\
Thuwal 23955, Saudi Arabia}
\and
\IEEEauthorblockN{Halima Bensmail$^*$}
\IEEEauthorblockA{Qatar Computing Research Institute\\
Doha 5825, Qatar\\
$^*$Corresponding author}
}

\IEEEoverridecommandlockouts
\IEEEpubid{\makebox[\columnwidth]{978-1-4799-0652-9/13/\$31.00~\copyright2013
IEEE \hfill} \hspace{\columnsep}\makebox[\columnwidth]{ }}

\maketitle
\thispagestyle{plain}

\fancypagestyle{plain}{
\fancyhf{}	
\fancyfoot[L]{}
\fancyfoot[C]{}
\fancyfoot[R]{}
\renewcommand{\headrulewidth}{0pt}
\renewcommand{\footrulewidth}{0pt}
}

\pagestyle{fancy}{
\fancyhf{}
\fancyfoot[R]{}}
\renewcommand{\headrulewidth}{0pt}
\renewcommand{\footrulewidth}{0pt}

\begin{abstract}
In this paper we study the problem of learning from multiple modal data for purpose of document classification. In this problem, each document is composed two different modals of data, i.e., an image and a text. Cross-modal factor analysis (CFA) has been proposed to project the two different modals of data to a shared data space, so that the classification of a image or a text can be performed directly in this space. A disadvantage of CFA is that it has ignored the supervision information. In this paper, we improve CFA by incorporating the supervision information to represent and classify both image and text modals of documents. We project both image and text data to a shared data space by factor analysis, and then train a class label predictor in the shared space to use the class label information. The factor analysis parameter and the predictor parameter are learned jointly by solving one single objective function. With this objective function, we minimize the distance between the projections of image and text of the same document, and the classification error of the projection measured by hinge loss function. The objective function is optimized by an alternate optimization strategy in an iterative algorithm. Experiments in two different multiple modal document data sets show the advantage of the proposed algorithm over other CFA methods.
\end{abstract}

\begin{IEEEkeywords}
Multiple modal learning,
Cross-modal factor analysis,
Supervised learning
\end{IEEEkeywords}

\IEEEpeerreviewmaketitle

\section{Introduction}

In this paper, we deal with the problem of learning from multiple modal data \cite{Carenzi2004525,Lumini20061706}. Traditional data representation, classification and retrieval problems usually focus on single modal data \cite{Lee2014291,Szymczyk2015354}. For example, for the problem of text classification, we usually only consider using a data set of text to train a classifier \cite{Merkl199861,Kim2005403}. While for the problem of image representation, only the images are considered to learn the representation parameters \cite{Caicedo201250,Hong201394}.  Zhang et al. developed and optimized association rule mining algorithms and implemented them on paralleled micro-architectural platforms \cite{zhang2013accelerating}, \cite{zhang2011gpapriori}. However,  in modern information landscape, the data is usually composed of different modals. For example, in video clips, there are two modals of data, e.g. image sequence data and audio data. Moreover, in a research article, there are not only the text data, but also the image data. Learning from these multiple modal data has attracted much attention from both machine learning and multimedia information processing communities \cite{6573933,13367863}. Recently, cross-modal factor analysis (CFA) has been proposed to project different modal of data to a shared feature space so that classification or retrieval can be performed cross data modal \cite{li2003multimedia}.  It assumes that each document is composed of two modals of data, e.g., image and text, and try to learn a projection matrix for each modal, so that the projections of the two modals of a document can be as close to each other as possible. With these projections, we can project any data of one of the two modals to the shared data space, and then perform retrieval cross these two modals. Moreover, since the data of different modals can be projected to a shared data space, we can also train a classifier from both the modals for the classification of data of any modal. CFA is further extended to its kernel version in \cite{wang2011kernel}.

\IEEEpubidadjcol

In this paper, we consider the problem of learning a classifier from multiple modal data, and using the classifier to classify a single modal data. In a training set of documents, each document is composed of data of two modals, e.g., an image and a text. We can first apply CFA to project two modals to a shared space and then learn a classifier in this space by using the class label information. However, in the phase of projection, the class label information has been ignored by CFA. Actually, without using supervision information provided by class label information, it cannot be guaranteed that the projections are discriminative enough. Although we can apply powerful classification methods after the projection of multiple modal data, the discriminative information which is necessary for classification may has been lost during the projection procedure. Thus it is very necessary to include the available class label information in the CFA projection. Surprisingly, no work has been done to incorporate the supervision information contained by the class labels to improve the discriminative ability of CFA. To fill this gap, in this paper, we proposed the first supervised CFA method by regularizing the projections of different modals by the class label information.

The contributions of this paper are of two folds:

\begin{enumerate}
\item For the first time, we propose the formulation of supervised CFA. We propose to project the data of two modals to a shared data space by orthonormal transformations, and use a linear class label function to predict the class labels of the training multiple modal documents. To formulate the problem, we propose to minimize the difference between the projections of two modals of the same document, and simultaneously minimize the classification error of both modals measured by hinge loss. In this way, the learning of orthonormal transformation matrices of multiple modal projections and the class label predictor parameter are unified, and the predictor learning can regularize the learning of orthonormal transformation matrices to improve the discriminative ability of the multiple modal projections.
\item We also develop an iterative algorithm to optimize the constrained minimization problem of this formulation. The projection parameter and the predictor parameter are optimized alternately in an iterative algorithm. The orthonormal transformation matrices are optimized by fixing class label predictor parameter matrix and solving a  singular value decomposition (SVD) problem \cite{Liu201475,Zhang201575}. The class label predictor parameter matrix is solved by fixing orthonormal transformation matrices and solving a quadratic programming (QP) problem. \cite{Miao201480,Fomeni2014173}
\end{enumerate}

The remaining of this paper is organized as follows: in section \ref{sec:method} we introduce the proposed supervised CFA (SupCFA), in section \ref{sec:experiment} the proposed method is evaluated experimentally, and in section \ref{sec:conclusion}, the paper is concluded.

\section{Proposed method}
\label{sec:method}

\subsection{Problem formulation}

We assume that we have a training set of $n$ documents $\mathcal{D}=\{D_1, \cdots, D_n\}$, where $D_i$ is the $i$-th document. Each document is comprised of an image component and an accompanying text, i.e., $D_i = (I_i, T_i)$, where $I_i \in \mathbb{R}^{d_I}$ is a $d_I$-dimensional feature space of the image, $T_i \in \mathbb{R}^{d_T}$ is a a $d_T$-dimensional feature of the text. These documents are assumed to belongs to $m$ classes, and for the $i$-th document, we define a class label vector $\bfy_i =[y_{i1},\cdots, y_{im}] \in \{+1,-1\}^m$ to indicate which class it belongs to. The $j$-th dimension of $\bfy_{i}$, $y_{ij} = +1$ if $D_i$ belongs to the $j$-th class, and $y_{ij} = -1$ otherwise. The goal of cross-modal classification is to learn a predictor from the training set, and use the predictor to predict the class label vector of given text (image) query $T$ ($I$).

To this end, we first project images and texts of documents to a shared $d$-dimensional feature space by orthonormal transformations, and then learn a linear prediction function in the share space to predict the class label vector. The projection in the image and text space are given as follows:

\begin{equation}
\begin{aligned}
I_i \rightarrow  I_i \Omega_I \in \mathbb{R}^d,~
T_i \rightarrow T_i \Omega_T  \in \mathbb{R}^d
\end{aligned}
\end{equation}
where $\Omega_I \in \mathbb{R}^{d_I \times d} $ is the orthonormal transformation matrix of the image data, and $\Omega_T \in \mathbb{R}^{d_T\times d} $ is the orthonormal transformation matrix of the text data. CFA assumes that the projections of the image and text of a single document should be as close to each other as possible, and the squared $\ell_2$ norm distance between $I_i \Omega_I$ and $T_i \Omega_T  $ is minimized over all the training documents,

\begin{equation}
\label{equ:CFA}
\begin{aligned}
\min_{\Omega_I, \Omega_T}
~&\sum_{i=1}^n \|\Omega_I I_i - \Omega_T T_i\|_2^2\\
s.t.
~& \Omega_I^\top \Omega_I = I_{d\times d}, \Omega_T^\top \Omega_T = I_{d\times d},
\end{aligned}
\end{equation}
where $I_{d\times d}$ is a $d_I\times d_I$ identity matrix. To predict the class label vector $\bfy_i$ from the projections of image and text of the $i$-th document, we try to learn a linear function as follows,

\begin{equation}
\begin{aligned}
&\bfy_i\leftarrow f_W(I_i \Omega_I ) =  (I_i\Omega_I ) W, \\
&\bfy_i\leftarrow  f_W(T_i\Omega_T ) =  (\Omega_T T_i) W,
\end{aligned}
\end{equation}
where $W\in \mathbb{R}^{d \times m }$ is the predictor parameter matrix. To learn $W$, we minimize its squared $\ell_2$ norm and the hinge loss of the predictor over both the images and texts of all the training documents,

\begin{equation}
\label{equ:SVM}
\begin{aligned}
\min_{W}
~&\|W\|_2^2 + C_1\sum_{i=1}^n (\xi_i + \varepsilon_i)\\
s.t.
~&h-  (I_i \Omega_I ) W \bfy_i^\top  \leq \xi_i, \xi_i\geq 0,\\
~&h- (\Omega_T T_i) W  \bfy_i^\top \leq \varepsilon_i, \varepsilon_i\geq 0, i=1,\cdots, n.\\
\end{aligned}
\end{equation}
where $\xi_i$ is the slack variable of the hinge loss over the image of the $i$-th document, $\varepsilon_i$ is that of the text of the $i$-th document, $h$ is a parameter of the hinge loss function, and $C_1$ is a tradeoff parameter. The minimization of $\|W\|_2^2$ is to reduce the complexity of the predictor and also to seek a large margin. The hinge loss is applied to reduce the prediction error.

The formulation of the proposed method is the combination of (\ref{equ:CFA}) and (\ref{equ:SVM}), which is as follows,

\begin{equation}
\label{equ:objective}
\begin{aligned}
\min_{W,\Omega_I, \Omega_T}
~&\frac{1}{2}\|W\|_2^2 + C_1\sum_{i=1}^n (\xi_i + \varepsilon_i) + C_2\sum_{i=1}^n \|I_i \Omega_I - T_i \Omega_T \|_2^2\\
s.t.
~&h-  (I_i \Omega_I ) W \bfy_i^\top  \leq \xi_i, \xi_i\geq 0,\\
~&h- (T_i \Omega_T ) W  \bfy_i^\top \leq \varepsilon_i, \varepsilon_i\geq 0, i=1,\cdots, n,\\
~&\Omega_I^\top \Omega_I = I_{d\times d}, \Omega_T^\top \Omega_T = I_{d\times d},
\end{aligned}
\end{equation}
where $C_2$ is another tradeoff parameter. Please note that in this formulation, not only the orthonormal transformation matrices are variables, but also the predictor parameter. In this way, the representation and the classification of image and text modals are unified. Both the images and texts are mapped to a shared space and then a shared predictor are applied to classify them.

\subsection{Optimization}

To solve the problem in (\ref{equ:objective}), we write the Lagrange function as follows,

\begin{equation}
\label{equ:Lagrange}
\begin{aligned}
\mathcal{L}
=&\frac{1}{2} \|W\|_2^2 + C_1\sum_{i=1}^n (\xi_i + \varepsilon_i) + C_2\sum_{i=1}^n \|I_i \Omega_I - T_i \Omega_T \|_2^2\\
&+\sum_{i=1}^n \alpha_i \left (h-  (I_i \Omega_I ) W \bfy_i^\top  - \xi_i \right ) - \sum_{i=1}^n \beta_i \xi_i\\
~&+\sum_{i=1}^n \gamma_i \left ( h- (T_i \Omega_T ) W  \bfy_i^\top - \varepsilon_i \right ) -\sum_{i=1}^n \delta_i \varepsilon_i\\
~&- Tr \left (\Gamma^\top \left (\Omega_I^\top \Omega_I - I_{d\times d} \right )\right )
-Tr \left (\Delta^\top \left (\Omega_T^\top \Omega_T - I_{d\times d}\right ) \right ),
\end{aligned}
\end{equation}
where $\alpha_i$, $\beta_i$, $\gamma_i$, $\delta_i, i=1,\cdots, n$, $\Gamma$ and $\Delta$ are Lagrange multiplier variables. To seek the minimization of the objective, we set the partial derivative of $\mathcal{L}$ with regard to $W$, $\xi_i$ and $\varepsilon_i$ to zero respectively,

\begin{equation}
\label{equ:partial}
\begin{aligned}
&\frac{\partial \mathcal{L}}{\partial W}
= W
-\sum_{i=1}^n \alpha_i  (I_i \Omega_I)^\top \bfy_i
-\sum_{i=1}^n \gamma_i (T_i \Omega_T )^\top \bfy_i = 0,\\
& \Rightarrow
W =
\sum_{i=1}^n \alpha_i  (I_i \Omega_I)^\top \bfy_i
+\sum_{i=1}^n \gamma_i (T_i \Omega_T )^\top \bfy_i,\\
&\frac{\partial \mathcal{L}}{\partial \xi_i}
= C_1 - \alpha_i  -  \beta_i = 0
\Rightarrow
C_1 - \alpha_i   = \beta_i \geq 0
\Rightarrow
\alpha_i  \leq C_1 \\
&\frac{\partial \mathcal{L}}{\partial \varepsilon_i}
= C_1 - \gamma_i  -  \delta_i = 0
\Rightarrow
C_1 - \gamma_i   = \delta_i \geq 0
\Rightarrow
\gamma_i  \leq C_1 .
\end{aligned}
\end{equation}
By substituting (\ref{equ:partial}) to (\ref{equ:Lagrange}), we have

\begin{equation}
\label{equ:Lagrange1}
\begin{aligned}
\mathcal{L}
=& f(\alpha_i|_{i=1}^n, \gamma_i|_{i=1}^n, \Omega_I,\Omega_T)+g(\alpha_i|_{i=1}^n, \gamma_i|_{i=1}^n, \Omega_I,\Omega_T, \Delta, \Gamma)
\end{aligned}
\end{equation}
where

\begin{equation}
\begin{aligned}
& f(\alpha_i|_{i=1}^n, \gamma_i|_{i=1}^n, \Omega_I,\Omega_T)\\
& =
-\frac{1}{2} \sum_{i,j=1}^n \alpha_i \alpha_j Tr\left (\Omega_I^\top I_i^\top \bfy_i \bfy_j^\top I_j \Omega_I \right )\\
&+
\sum_{i,j=1}^n \alpha_i \gamma_j Tr \left (\Omega_I I_i \bfy_i ^\top \bfy_j T_j  \Omega_T  \right )\\
&
-
\frac{1}{2} \sum_{i,j=1}^n \gamma_i \gamma_j Tr\left (\Omega_T^\top T_i^\top \bfy_i \bfy_j^\top T_j  \Omega_T  \right )\\
&+C_2 \sum_{i=1}^n Tr \left (\Omega_I^\top I_i ^\top I_i \Omega_I \right ) - 2 C_2  \sum_{i=1}^n Tr \left (\Omega_I^\top I_i ^\top T_i \Omega_T  \right )+\\
& C_2 \sum_{i=1}^n Tr \left (\Omega_T^\top T_i ^\top T_i \Omega_T  \right ), ~and\\
&g(\alpha_i|_{i=1}^n, \gamma_i|_{i=1}^n, \Omega_I,\Omega_T, \Delta, \Gamma)\\
&= h \sum_{i=1}^n ( \alpha_i + \gamma_i ) \\
&- Tr \left (\Gamma^\top \left (\Omega_I^\top \Omega_I - I_{d\times d} \right )\right )
-Tr \left (\Delta^\top \left (\Omega_T^\top \Omega_T - I_{d\times d}\right ) \right ).
\end{aligned}
\end{equation}

The optimization problem is transferred to the following coupled problem,

\begin{equation}
\label{equ:Lagrange2}
\begin{aligned}
\max_{\alpha_i|_{i=1}^n, \gamma_i|_{i=1}^n, \Gamma, \Delta}~ &\min_{\Omega_I, \Omega_T}
~\mathcal{L},\\
s.t.
~& 0 \leq \alpha_i \leq C_1, 0 \leq \gamma_i \leq C_1, i=1,\cdots, n,\\
&\Gamma \geq 0, \Delta \geq 0.
\end{aligned}
\end{equation}
To solve this problem, we employ the alternate optimization strategy. We optimize $\alpha_i|_{i=1}^n, \gamma_i|_{i=1}^n$ and $\Gamma, \Delta, \Omega_I, \Omega_T$ alternately in an iterative algorithm. In each iteration, we first fix $\Gamma, \Delta, \Omega_I, \Omega_T$ and optimize $\alpha_i|_{i=1}^n, \gamma_i|_{i=1}^n$, and then we fix $\alpha_i|_{i=1}^n, \gamma_i|_{i=1}^n$ and optimize $\Gamma, \Delta, \Omega_I, \Omega_T$.

\subsubsection{Optimizing $\alpha_i|_{i=1}^n$ and $\gamma_i|_{i=1}^n$}

Fixing $\Gamma, \Delta, \Omega_I$ and $\Omega_T$, and removing the terms irrelevant to $\alpha_i|_{i=1}^n$ and $\gamma_i|_{i=1}^n$ from (\ref{equ:Lagrange1}), we rewrite (\ref{equ:Lagrange2}) as

\begin{equation}
\label{equ:Lagrange3}
\begin{aligned}
\max_{\alpha_i|_{i=1}^n, \gamma_i|_{i=1}^n}
~& -\frac{1}{2} \sum_{i,j=1}^n \alpha_i \alpha_j Tr\left (\Omega_I^\top I_i^\top \bfy_i \bfy_j^\top I_j \Omega_I \right )\\
&
+
\sum_{i,j=1}^n \alpha_i \gamma_j Tr \left (\Omega_I I_i \bfy_i ^\top \bfy_j T_j  \Omega_T  \right )\\
&
-
\frac{1}{2} \sum_{i,j=1}^n \gamma_i \gamma_j Tr\left (\Omega_T^\top T_i^\top \bfy_i \bfy_j^\top T_j  \Omega_T  \right )\\
&+h \sum_{i=1}^n ( \alpha_i + \gamma_i )\\
s.t.
~& 0 \leq \alpha_i \leq C_1, 0 \leq \gamma_i \leq C_1, i=1,\cdots, n.
\end{aligned}
\end{equation}
This problem can be solved as a QP problem.

\subsubsection{Optimizing $\Omega_I$ and $\Omega_T$}

By fixing $\alpha_i|_{i=1}^n$ and  $\gamma_i|_{i=1}^n$, and removing terms irrelevant to $\Gamma, \Delta, \Omega_I$ and $\Omega_T$ from (\ref{equ:Lagrange1}), we rewrite (\ref{equ:Lagrange2}) as

\begin{equation}
\label{equ:Omega}
\begin{aligned}
\max_{\Gamma, \Delta}~ \min_{\Omega_I, \Omega_T}
~
& -\frac{1}{2} \sum_{i,j=1}^n \alpha_i \alpha_j Tr\left (\Omega_I^\top I_i^\top \bfy_i \bfy_j^\top I_j \Omega_I \right )\\
&
+
\sum_{i,j=1}^n \alpha_i \gamma_j Tr \left (\Omega_I^\top I_i^\top \bfy_i  \bfy_j^\top T_j  \Omega_T \right )\\
&
-
\frac{1}{2} \sum_{i,j=1}^n \gamma_i \gamma_j Tr\left (\Omega_T^\top T_i^\top \bfy_i \bfy_j^\top T_j  \Omega_T  \right )\\
&+C_2 \sum_{i=1}^n Tr \left (\Omega_I^\top I_i ^\top I_i \Omega_I \right ) \\
&- 2 C_2  \sum_{i=1}^n Tr \left (\Omega_I^\top I_i ^\top T_i \Omega_T  \right )\\
& + \sum_{i=1}^n Tr \left (\Omega_T^\top T_i ^\top T_i \Omega_T  \right )\\
&- Tr \left (\Gamma^\top \left (\Omega_I^\top \Omega_I - I_{d\times d} \right )\right )\\
&
-Tr \left (\Delta^\top \left (\Omega_T^\top \Omega_T - I_{d\times d}\right ) \right ),\\
s.t.
~& \Gamma \geq 0, \Delta \geq 0.
\end{aligned}
\end{equation}
The primal problem of this dual problem is

\begin{equation}
\label{equ:primal}
\begin{aligned}
\min_{\Omega_I, \Omega_T}
~
& -\frac{1}{2} \sum_{i,j=1}^n \alpha_i \alpha_j Tr\left (\Omega_I^\top I_i^\top \bfy_i \bfy_j^\top I_j \Omega_I \right )\\
&
+
\sum_{i,j=1}^n \alpha_i \gamma_j Tr \left (\Omega_I^\top I_i^\top \bfy_i  \bfy_j^\top T_j  \Omega_T \right )\\
&
-
\frac{1}{2} \sum_{i,j=1}^n \gamma_i \gamma_j Tr\left (\Omega_T^\top T_i^\top \bfy_i \bfy_j^\top T_j  \Omega_T  \right )\\
&+C_2 \sum_{i=1}^n Tr \left (\Omega_I^\top I_i ^\top I_i \Omega_I \right ) - 2 C_2  \sum_{i=1}^n Tr \left (\Omega_I^\top I_i ^\top T_i \Omega_T  \right ) \\
&+ \sum_{i=1}^n Tr \left (\Omega_T^\top T_i ^\top T_i \Omega_T  \right )\\
s.t.
~& \Omega_I^\top \Omega_I = I_{d\times d},
\Omega_T^\top \Omega_T = I_{d\times d}.
\end{aligned}
\end{equation}
Using the constrains $\Omega_I^\top \Omega_I = I_{d\times d}$ and $\Omega_T^\top \Omega_T - I_{d\times d}$, we can remove some constant terms from (\ref{equ:primal}) and rewrite it as

\begin{equation}
\label{equ:primal1}
\begin{aligned}
\min_{\Omega_I, \Omega_T}
~
&
\sum_{i,j=1}^n \alpha_i \gamma_j Tr \left (\Omega_I^\top I_i^\top \bfy_i  \bfy_j^\top T_j  \Omega_T \right )\\
&
- 2 C_2  \sum_{i=1}^n Tr \left (\Omega_I^\top I_i ^\top T_i \Omega_T  \right ) \\
&=
Tr \left ( \Omega_I^\top \left (
\sum_{i,j=1}^n \alpha_i \gamma_j I_i^\top  \bfy_i \bfy_j^\top  T_j- 2 C_2  \sum_{i=1}^n  I_i ^\top T_i  \right ) \Omega_T \right ) \\
s.t.
~& \Omega_I^\top \Omega_I = I_{d\times d},
\Omega_T^\top \Omega_T = I_{d\times d}.
\end{aligned}
\end{equation}
The equivalent problem is

\begin{equation}
\label{equ:primal2}
\begin{aligned}
\max_{\Omega_I, \Omega_T}
~&
Tr \left ( \Omega_I^\top Z \Omega_T   \right ) \\
s.t.
~& \Omega_I^\top \Omega_I = I_{d\times d},
\Omega_T^\top \Omega_T = I_{d\times d}.
\end{aligned}
\end{equation}
where

\begin{equation}
\label{equ:Z}
\begin{aligned}
Z = \sum_{i,j=1}^n \alpha_i \gamma_j I_i^\top  \bfy_i \bfy_j^\top  T_j- 2 C_2  \sum_{i=1}^n  I_i ^\top T_i \in \mathbb{R}^{d_I \times d_T}.
\end{aligned}
\end{equation}
The optimal matrices $\Omega_I $ and $\Omega_T$ can be obtained by a
SVD of the matrix $Z$, i.e.,

\begin{equation}
\label{equ:primal5}
\begin{aligned}
Z=\Omega_I \Sigma \Omega_T^\top
\end{aligned}
\end{equation}
where $\Sigma$ is the matrix of singular values of $Z$, and $\Sigma$ is diagonal.

\subsection{Algorithm}

Based on the optimization results, we can design an iterative algorithm as Algorithm \ref{alg:iter}. The iterations are repeated $T$ times, and the $t$-th each iteration, the variables $\Omega_I^{t}$, $\Omega_T^{t}$, $\alpha_i^t|_{i=1}^n$ and  $\gamma_i^t|_{i=1}^n$ are updated alternately. Finally the predictor parameter $W$ is calculated from the updated variables.

\begin{algorithm}[htb!]
\caption{Iterative learning algorithm of supervised cross-modal factor analysis.}
\label{alg:iter}
\begin{algorithmic}
\STATE \textbf{Input}: A training set of $n$ documents $\{(I_1,T_i),\cdots,(I_n,T_n)\}$, and corresponding class label vector set $\{\bfy_1,\cdots,\bfy_n\}$;

\STATE \textbf{Input}: Tradeoff parameters $C_1$, $C_2$, and maximum iteration number $T$;

\STATE Initialize orthonormal transformation matrices $\Omega_I^0$ and $\Omega_T^0$;

\FOR{$t=1,\cdots,T$}
\STATE Fix $\Omega_I^{t-1}$ and $\Omega_T^{t-1}$, and update $\alpha_i^t|_{i=1}^n$ and  $\gamma_i^t|_{i=1}^n$ by solving the problem in (\ref{equ:Lagrange3});

\STATE Fix $\alpha_i^t|_{i=1}^n$ and $\gamma_i^t|_{i=1}^n$ , and update $Z^t$ as in (\ref{equ:Z});

\STATE Update $\Omega_I^{t}$ and $\Omega_T^{t}$ by applying SVD to $Z^t$ as in (\ref{equ:primal5});

\ENDFOR

\STATE Calculate predictor function parameter matrix $W$ from $\Omega_I^T$, $\Omega_T^T$, $\alpha_i^T|_{i=1}^n$ and $\gamma_i^T|_{i=1}^n$ (\ref{equ:partial}).

\STATE \textbf{Output}:
The orthonormal transformation matrices $\Omega_I^T$ and $\Omega_T^T$ and predictor function parameter matrix $W$.

\end{algorithmic}
\end{algorithm}

\section{Experiments}
\label{sec:experiment}

In this section, we will investigate the proposed algorithm experimentally.

\subsection{Experiment setup}

\subsubsection{Data sets}

In this experiment, we used two different document data sets composed of images and texts.
The first data set is the TVGraz database \cite{khan2009tvgraz}, which is a multimodal database of object categories composed of textual and visual features. The documents belongs to 10 of 256 classes of Caltech-256. 1,000 webpages are retrieved for each of the 10 classes, and 2,058 image-text pairs are collected. Each image-text pair is linked to a document, thus we have 2,058 documents of 10 classes in this data set.

The seconde data set is the Wikipedia database, which is selected from Wikipedia featured article database, and Wikipedia featured article database contains documents of 30 classes. Because most of the classes of Wikipedia featured articles database contains very few documents, we only choose the 10 classes with the most documents. Moreover, each featured article usually have more than one image and section, so we split each featured article to several documents. Each document contains a section of a featured article, and the images placed to this section. We have in total 7,114 documents. Moreover, we remove the documents which have more than one image, and the documents which have a text with less than 70 words. Finally, we have 2,866 documents in total in our experiment.

The images in the documents are represented as feature vectors using the bag-of-features method, while the texts in the documents are represented as feature vectors using the bag-of-words method.

\subsubsection{Experiment protocol}

To conduct the experiment, we used the 10-fold cross validation strategy. The entire data set is split to 10 folds randomly, and each fold was used as a test set in turn, while the remaining 9 sets were combined and used as a training set. The proposed learning algorithm was performed to the training set to learn the orthonormal transformation matrices and the predictor parameter matrix, and then they are used to represent and classify the individual images and texts in the test set. The classification accuracy is measured by the classification rate as follows,

\begin{equation}
\label{equ:rate}
\begin{aligned}
&classification~ rate \\
&= \frac{number~of~correctly ~classified~images~and~texts}{total~number~of~test~images~and~texts}.
\end{aligned}
\end{equation}

\subsection{Results}

\subsubsection{Comparison to unsupervised CFA}

\begin{figure}[!htb]
  \centering
\subfigure[TVGraz]{
\includegraphics[width=0.23\textwidth]{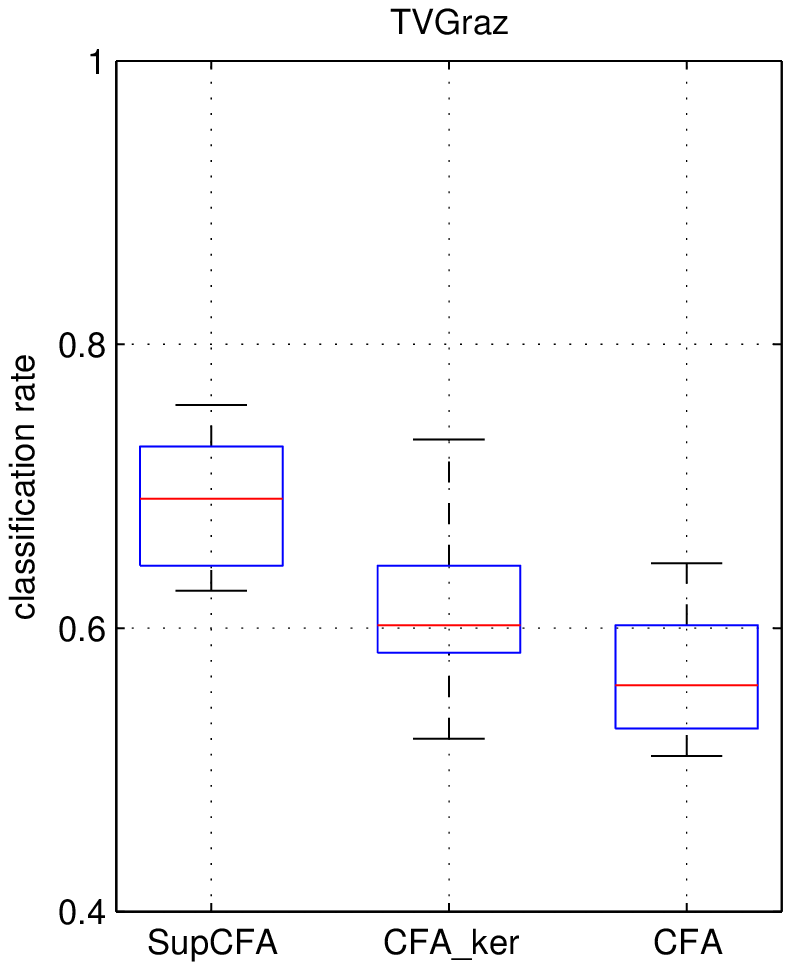}}
\subfigure[Wikipedia]{
\includegraphics[width=0.23\textwidth]{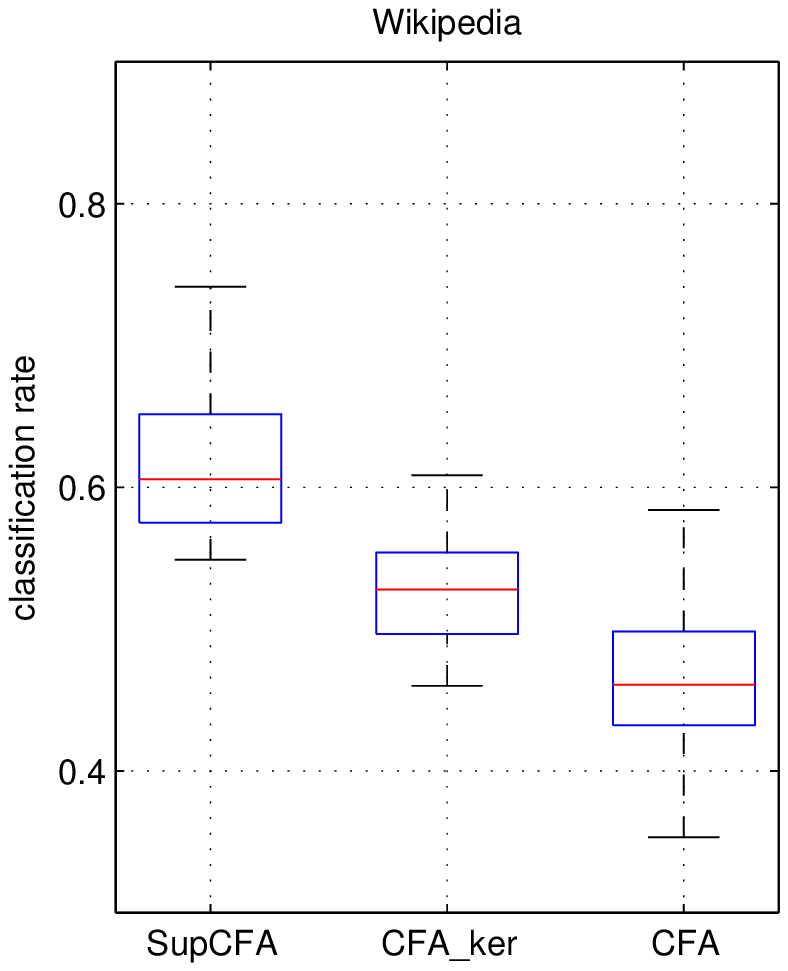}}
\\
\caption{Boxplots of classification rates of 10-fold cross validation of different CFA methods.}
\label{fig:comparision}
\end{figure}

We first compare the proposed supervised version of CFA against unsupervised CFA methods on the problem of image/text classification problem. We considered the original CFA method \cite{li2003multimedia} and its kernel version (CFA$_{ker}$) \cite{wang2011kernel} as data representation methods, and used a SVM as a classifier. The boxplots of classification rates of the 10-fold cross validations of the compared methods over two data sets are given in Fig. \ref{fig:comparision}. It is clear that the proposed SupCFA outperforms the two unsupervised CFA methods completely. The low quartile of the SupCFA classification rates is even higher than the upper quartiles of the tow compared methods. This is not surprising at all because SupCFA is the only method which can explore the supervision information to improve the discriminative ability of cross-modal factor analysis, while CFA and CFA$_{ker}$ ignore the class label information at all. Moreover, it seems that CFA$_{ker}$  outperforms CFA due to its usage of kernel tricks.

\subsubsection{Algorithm convergency}

\begin{figure}
  \centering
  \includegraphics[width=0.3\textwidth]{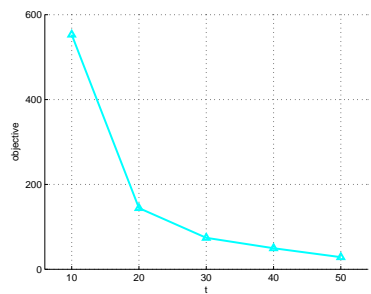}\\
  \caption{Convergency curve.}
  \label{fig:conver}
\end{figure}

Since the proposed algorithm ia an iterative algorithm, it is important to study its convergency over iterations. We plot the curve of objective function over different iterations in Fig. \ref{fig:conver}. It could be seen that the objective function is reduced significantly in the first 30 iterations, and it tends to converge after the 30-th iteration.

\section{Conclusions and Future works}
\label{sec:conclusion}

In this paper, we proposed the first supervised CFA method for the presentation and classification of multiple modal data. The proposed method not only projects data of different modals to a shared data space like CFA, but also tries to learn a predictor to predict the class labels from this data space. Moreover, the learning of projection and class label prediction parameters are learned within a single objective function. By optimizing this objective function with regard to both projection and class label prediction parameters jointly, the class label information is used to guild the learning of CFA parameters. The experiment results show that the supervised CFA outperforms both linear and kernel versions of CFA without considering class label information. 
We also plan to explore potential of proposed algorithm using big data analysis and high performance computing
\cite{wang2015towards,zhang2014lucas,gao2014sparse,zhang2013fpga,zhang2011frequent,zhang2013accelerating,zhang2011gpapriori,li2013zht,zhao2014fusionfs,li2013distributed,wangusing,wang2014optimizing,wang2014next,wangovercoming,wang2015towards},
for applications of big multimodal data processing \cite{6984567,wang2012scimate,wang2013supporting,wang2014saga,wang2015maximum,wang2014removing},
information and network security \cite{Xu2014TAAS,Xu2012TAAS,xu2014evasion,zhan2013characterizing,xu2013cross,sun2012unsupervised,sun2014primate,sun2013multi,sun2014human,sun2014mobile}, computer vision \cite{wang2013joint,wang2013multiple,wang2014beyond,sun2015non,wang2014feature,wang2012adaptive,wang2013discriminative,luo2011piecewise,luo2011piecewise1},
and bioinformatics \cite{wang2012mathematical,wang2010conceptual,yang2012mathematical,hu2009improving,zhang2010bioinformatics,zhang2009bayesian,hu2009improving,Bhuyan2011,wang2012multiple,liu2013structure,zhou2014biomarker,wang2013non,wang2012proclusensem,wang2012prodis,dai2012bioinformatics,peng2015modeling,wang2014computational}.

\section*{Acknowledgements}

The research reported in this publication was supported by competitive research funding from King Abdullah University of Science and Technology (KAUST), Saudi Arabia.



\end{document}